# Lung Cancer Detection using Co-learning from Chest CT Images and Clinical Demographics

Jiachen Wang [a], Riqiang Gao [a], Yuankai Huo *[b], Shunxing Bao [a], Yunxi Xiong [a], Sanja L. Antic [c], Travis J. Osterman [d], Pierre P. Massion [c], Bennett A. Landman[a,b]

[a] Computer Science, Vanderbilt University, Nashville, TN, USA 37235
[b] Electrical Engineering, Vanderbilt University, Nashville, TN, USA 37235
[c] Division of Pulmonary and Critical Care Medicine, Department of Medicine, Vanderbilt University Medical Center, Vanderbilt Ingram Cancer Center, Nashville, TN, USA 37235
[d] Department of Biomedical Informatics, Vanderbilt University Medical Center, Nashville, TN, USA 37235

(*Corresponding Author: yuankai.huo@vanderbilt.edu)

## ABSTRACT

Early detection of lung cancer is essential in reducing mortality. Recent studies have demonstrated the clinical utility of low-dose computed tomography (CT) to detect lung cancer among individuals selected based on very limited clinical information. However, this strategy yields high false positive rates, which can lead to unnecessary and potentially harmful procedures. To address such challenges, we established a pipeline that co-learns from detailed clinical demographics and 3D CT images. Toward this end, we leveraged data from the Consortium for Molecular and Cellular Characterization of Screen-Detected Lesions (MCL), which focuses on early detection of lung cancer. A 3D attention-based deep convolutional neural net (DCNN) is proposed to identify lung cancer from the chest CT scan without prior anatomical location of the suspicious nodule. To improve upon the non-invasive discrimination between benign and malignant, we applied a random forest classifier to a dataset integrating clinical information to imaging data. The results show that the AUC obtained from clinical demographics alone was 0.635 while the attention network alone reached an accuracy of 0.687. In contrast when applying our proposed pipeline integrating clinical and imaging variables, we reached an AUC of 0.787 on the testing dataset. The proposed network both efficiently captures anatomical information for classification and also generates attention maps that explain the features that drive performance.

## 1. INTRODUCTION

The American Cancer Society estimates that there is an increasing trend on the number of patients that are diagnosed with lung cancer in the US [1]. Early detection of lung cancer is essential in reducing mortality, as the survival rate decreases dramatically in later stages of lung cancer compared to earlier ones. Researchers have conducted collaborative effort in multi-site large-scale cohorts, for example the National Lung Screening Trial (NLST) [2]. Historically, such cohorts have been image-centered with limited clinical information and such approaches may yield high false positive rates, which can lead to unnecessary diagnostic procedures [3]. To address such challenges, the NCI sponsored a Consortium for Molecular and Cellular Characterization of Screen-Detected Lesions to which we are members of. The MCL focuses on detection of early stage lung cancer and captures not only CT images, but also comprehensive clinical information, which offers unique opportunities to decrease overdiagnosis in early stage lung cancer.

In this project, we address the problem of discrimination between benign and malignant nodules on screening and non screening-scans (Figure 1). We propose a machine learning based lung cancer benign/malignant classification pipeline, which co-learns with 3D CT image volumes and clinical demographics. The proposed method integrates deep convolutional neural networks (DCNN) and other statistical and machine learning techniques to characterize early-stage lung cancer and distinguish benign from malignant nodules (Figure 2). To visualize the behavior of the deep neural network, the 3D attention mechanism [5] was integrated to the DCNN. This proposed method achieved superior performance on classification accuracy by co-learning with image features and clinical demographics than learning from a single resource respectively.

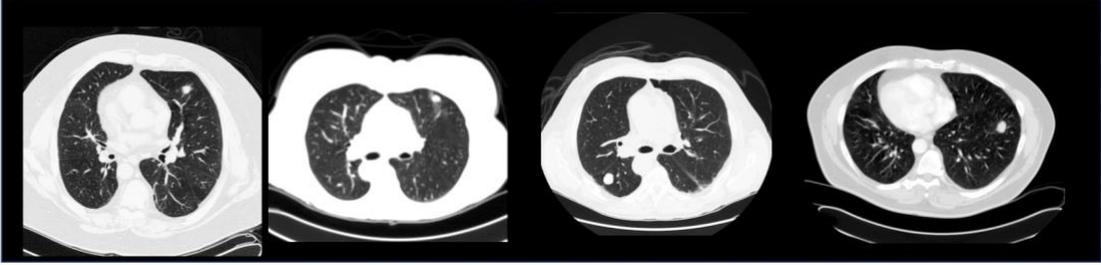

Figure 1. Problem to address: our initiative to render benign/malignant classification for lung cancer. In the above table shows ground truths, predictions from image-only (CNN) method, clinical-only (RF) method and the proposed integrated pipeline. The proposed pipeline is able to make correct decisions in all the above four sample cases.

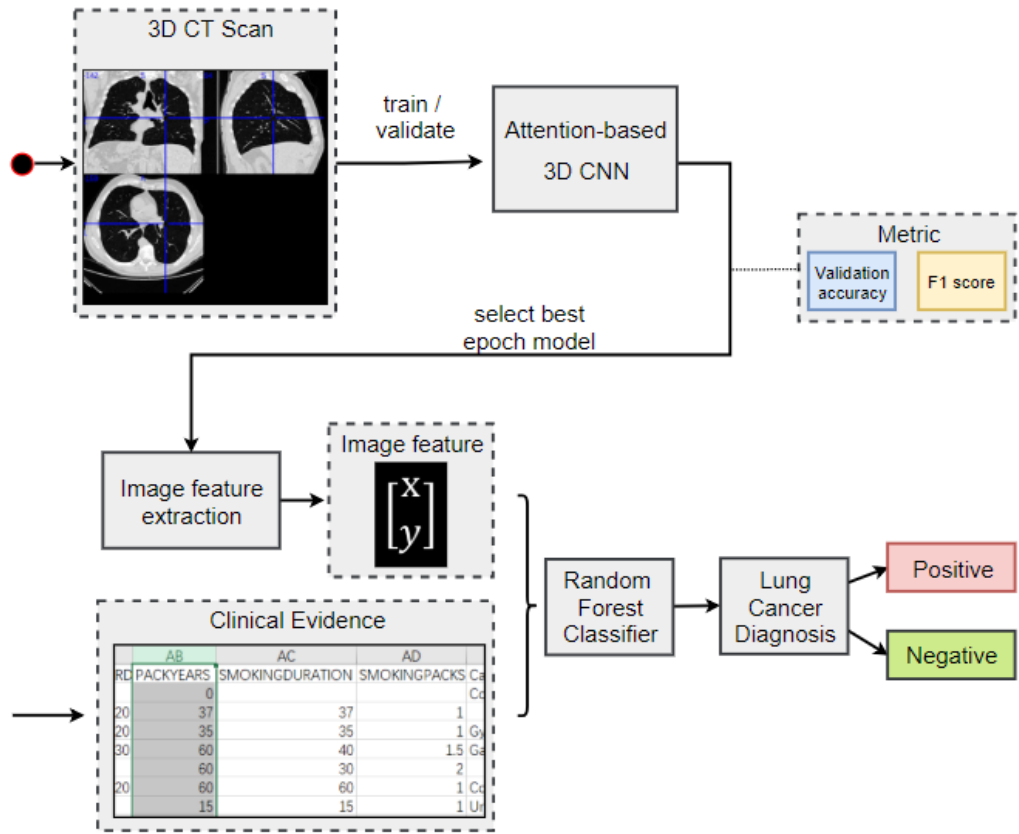

Figure 2. Method pipeline: the proposed pipeline co-learns with 3D CT images and clinical demographic information to classify early-stage lung cancer and predict biopsy determined diagnosis. This pipeline will be compared to image-only method, clinical-information-only method and Kaggle Top1 solution.

## 2. METHODS

### 2.1 Preprocessing

A preprocessing pipeline is deployed for all input scans. First, the intensities of 3D volume were resampled to 1mm×1mm×1mm isotropic resolution. Then, the lung segmentations (https://github.com/lfz/DSB2017) were computed from the original CT volume and we zero-padded the non-lung regions to Hounsfield unit score of 170, and these were normalized to 0-1 scale intensities. The preprocessed whole lung scan was used as the first input channel for learning. To guide the learning, a lung nodule mask is obtained through a pre-trained 3D CNN U-Net nodule detection network developed by the first-place team of Kaggle Data Science Bowl 2017 (referred as "Kaggle Top1") [4]. Next, the lung 3D binary nodule mask, with 1s on regions with nodules and 0s elsewhere, is used as the second channel of input. The two channels were then co-rescaled to a 4D tensor (the first dimension is number of channels) of shape 2×128×128×128 as the final input to the attention-based CNN. The dimension of 128 was decided by the memory limitation from the GPU. The learning target for the network is a binary category, with 1 indicating malignant and 0 indicating benign. However, there were multiple sub-categories beyond this broad binary classification, and the inner-class disparity can be as dramatic as the inter-class disparity. For example, malignant cases can involve Adenocarcinoma, Non-Small Cell Lung Cancer non otherwise specified (NSCLC), Small Cell Carcinoma and Squamous Cell Carcinoma. In our experiments, data augmentations were performed to leverage the learning procedures. The 3D volumes were randomly rotated on the xy-plane by a maximum of 10 degrees and translated by a maximum of 4 voxels in three dimensions.

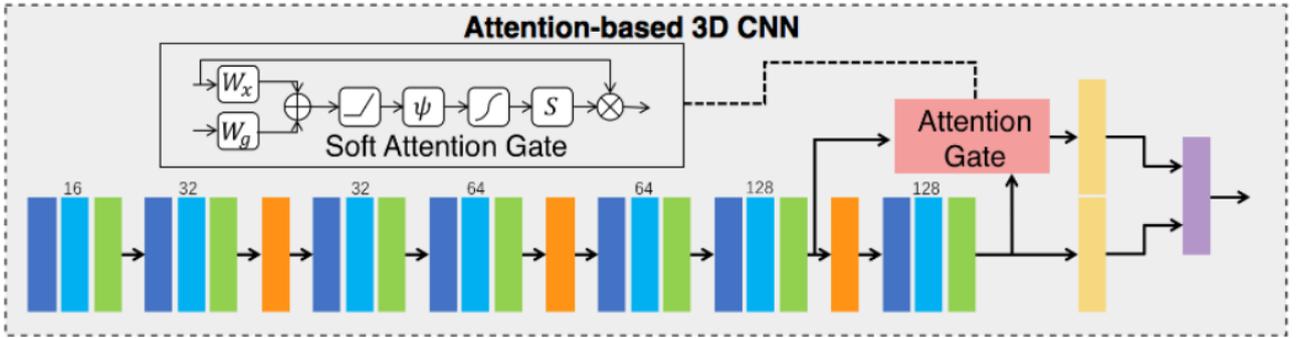

Figure 3. The 3D attention-based network structure, where every dark blue block is a Conv3D, every light blue block is a BatchNorm3D, every green block is a ReLU activation, and the three orange blocks are MaxPooling3D, and the two yellow blocks are AveragePooling3D, while the final purple layer is the dense layer to reduce the feature dimension to two. Structure of the soft attention gate (SAG) is also shown. The SAG filtered higher-level features with the attention gate from the lower level feature maps, and then the attention features were concatenated with main stream features for output.

Table 1: Demographic Information

|  | Training set | Validation set | Test Set |
|---|---|---|---|
| Samples size | 165 | 57 | 55 |
| Gender (Male, n %) | 101(61) | 32(56) | 34(61) |
| BMI (mean, std) | 27.512(6.512) | 26.511(5.251) | 29.010(6.675) |
| Age Quit smoking (mean, std) | 54.681(14.447) | 54.800(11.068) | 54.059(12.238) |
| Age Started smoking (mean, std) | 19.377(7.502) | 17.848(4.899) | 17.340(6.203) |
| CigsPerDay (mean, std) | 26.917(14.230) | 29.776(16.232) | 29.904(17.527) |
| CIGARETTE_Q1 (Current smoker, Ex-smoker, Never smoked) | (40, 114, 11) | (17, 37, 3) | (16, 36, 3) |
| Pack Years (mean, std) | 52.168(32.261) | 54.198(31.087) | 57.485(34.467) |
| Smoking Duration | 37.307(13.554) | 37.346(12.277) | 39.038(13.699) |
| Histologic Type: Benign | 40 | 13 | 10 |
| Histologic Type: Malignant | 125 | 44 | 45 |

## 2.2 Attention-based CNN

An attention-based CNN is proposed to train and validate on the 3D CT image scans. The 2D soft attention gate (SAG) [5] was extended to 3D in the proposed 3D network. The 3D convolutional layers, batch-normalization layers, max-pooling layers, average-pooling layers, and dense layers are used (Figure 3). The SAG filtered higher-level features with the attention gate from the lower level feature maps. Then, the attention features were concatenated with main stream features for final output.

## 2.3 Platform and Parameters

This project was implemented in Anaconda Python 3.6, and the deep learning framework adapted was Pytorch 0.4. Experiments were trained on a Nvidia Titan XP GPU that allowed a maximum batch size of 4 based on our model and input size selection. In the best-performing model, we used the traditional Adam optimizer with a modest learning rate of 0.0001 to smoothen the learning curve, and a traditional cost function of categorical cross-entropy loss is chosen.

## 2.4 Random Forest Classifier on Multi-source Data

A recently proposed random forest (RF) classifier, called "Xgboost" [6], has been used to learn from (1) clinical features only and, (2) both image and clinical features. For clinical features, BMI, cigarettes consumption per day, age starting and quitting smoking, smoking history, smoking packs per year, and gender are available and selected. For co-learning from clinical features and image features, the same clinical features from "clinical only" and the two-dimensional features from the final soft-max layer were used as multi-source features. Then, the multi-source features are fed to the RF classifier, with the exact same train-validation-split as the image and clinical only models. The feature importance map after training for both using clinical information only and using the entire pipeline, is shown in Figure 5 as quantitative results.

## 2.5. Population and study design.

In the MCL dataset, we captured 798 subjects (patients) in the XNAT radiology database at Vanderbilt University [7]. In the entire cohort, 387 subjects have both clinical information and chest CT image that had completed quality assurance (QA). Then, secondary QA was performed on both images and clinical demographic data to filter out the unusable CT scans and the subjects without valid clinical information or diagnosis other than [Adenocarcinoma, Granuloma, Large Cell Neuroendocrine, Negative for Malignant Cells, NSCLC, Normal, Small Cell Carcinoma, Squamous Cell Carcinoma].

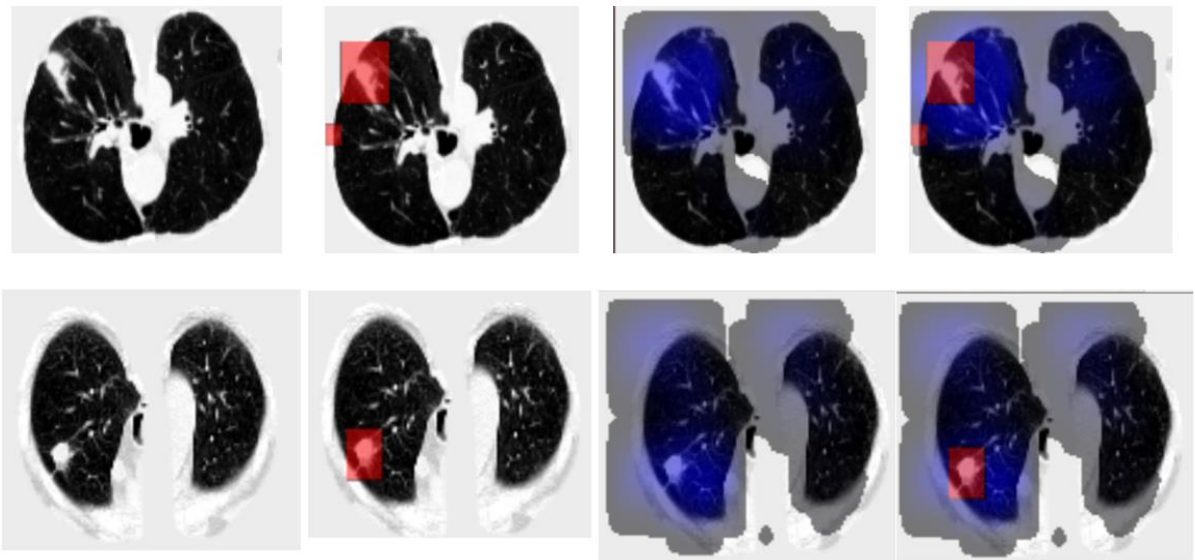

Figure 4: Qualitative results of two examples to show network attentions. The red boxes are detected nodule bounding boxes from Kaggle Top1 method [4] and the blue regions are testing attention mappings. It shows that the attention network is looking at regions that are highly indicative of potential lung cancer.

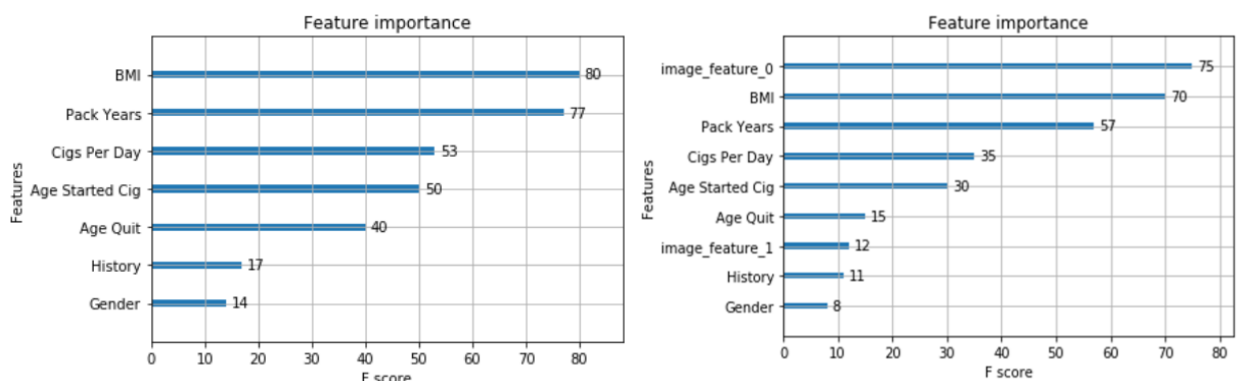

Figure 5: Quantitative results with feature importance plots. A higher score indicates greater importance. The left feature importance map is obtained by only using clinical demographic information, while the right is obtained through our proposed pipeline, with image_feature_0 and image_feature_1 being high-level features obtained from the attention network, where image_feature_0 and image_feature_1 are first and second dimension of the image feature from network output.

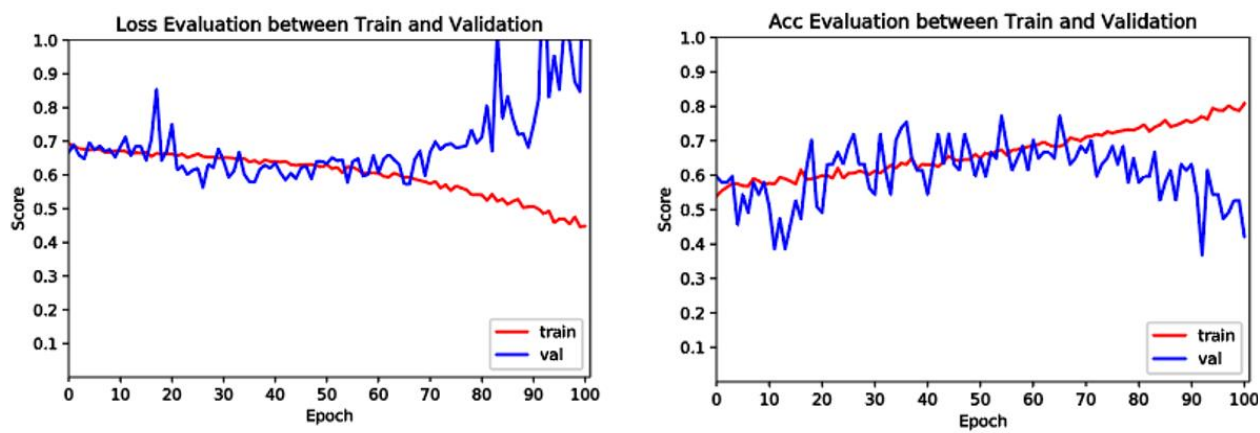

Figure 6. 3D attention-based network performance: On the left are the cross-entropy loss (Loss) curves between train and validation, and on the right are the accuracy (Acc) score curves between train and test on 100 epochs. The model's performance gradually improves until at around the 70$^{th}$ epoch when the model starts to overfit.

After QA, 277 subjects in the MCL were used in the experiments. The dataset was randomly divided into a training/validation cohort (80%) and a testing cohort (20%) for four-fold cross-validation within the training/validation cohort. The validation dataset provides an unbiased evaluation of a model fit on the training dataset while refining the model's variables. Note that the testing cohort never participated in cross-validation and was withheld from all training. For each fold, we had a split of ~167 training subjects (~60%), ~57 validation subjects (~20%), and the same fixed 55 withheld testing subjects (~20%). The detailed demographic information has been provided in Table 1. To improve the learning performance, we also used 1390 scans from Kaggle (including 358 positive samples and 1032 negative samples) National Data Science Bowl 2017 for only training purpose since the Kaggle dataset has the benign/malignant classes that are same as our learning purpose. The Kaggle dataset is combined with the training subjects from MCL dataset during training but does not participate in the validation or testing phase to avoid unnecessary bias. Note that all image scans don't have clinically annotated lung nodules.

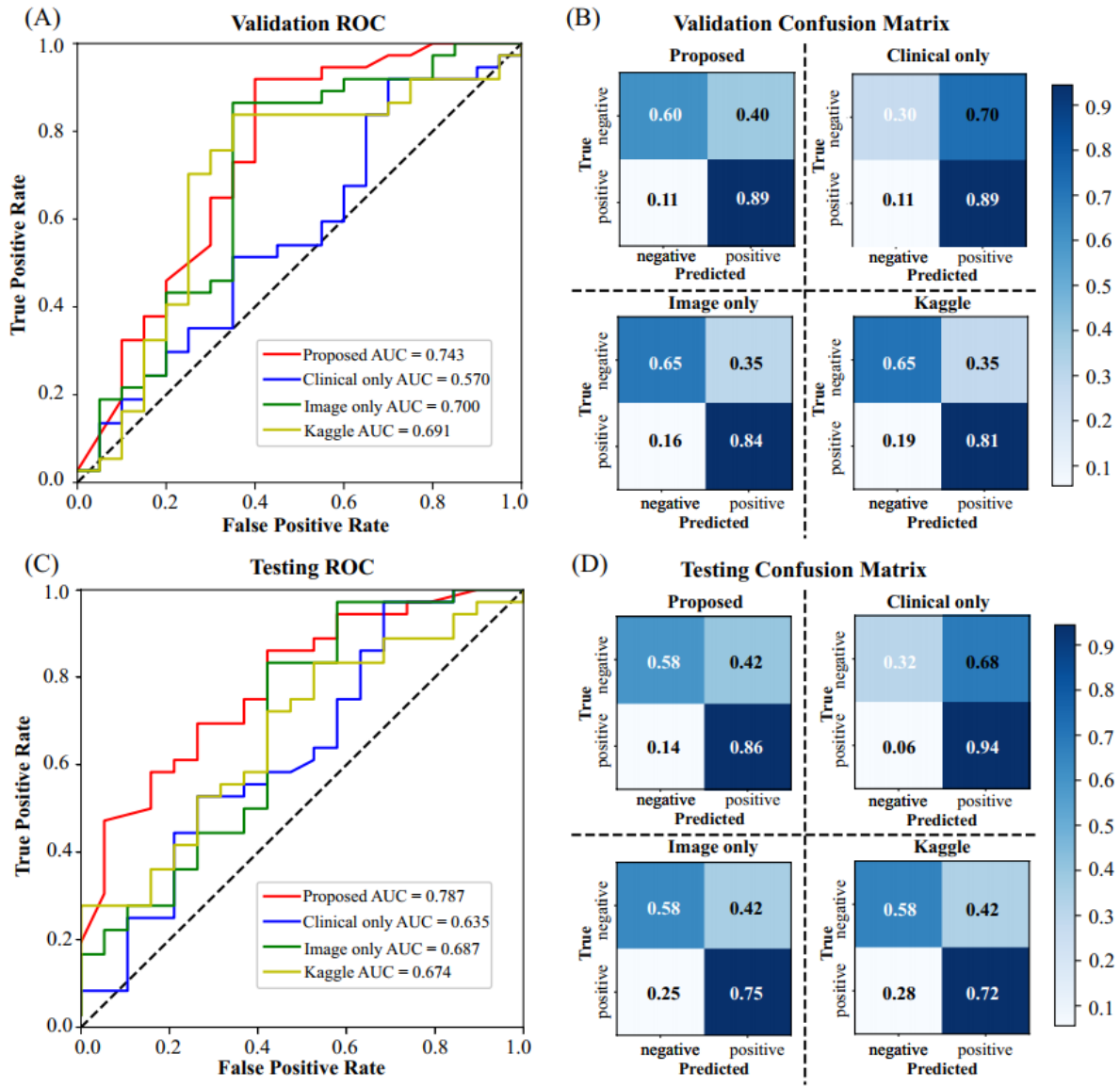

Figure 7. Quantitative results comparisons for the proposed pipeline, image-only method, using clinical-demographic-only method and Kaggle Top1 solution: (A) Validation ROC curves. (B) Validation confusion matrices. (C) Testing ROC curves. (D) Testing confusion matrices. The confusion matrices are plotted with class prediction thresholds selected based on ROC curves to render optimal results.

## 3.  RESULTS

The hyper-parameters for both DCNN and RF were tuned based on the validation scans to achieve the best validation performance. Then, the hyper-parameters and the trained models were deployed on the withhold testing data directly without further tuning to avoid the bias. Figure 6 shows the training and validation loss and accuracy (Acc) across the 100 training epochs. Figure 4 above shows the qualitative results from attention maps and our input lesion masks.

Figure 7 plots the Receiver Operating Characteristic (ROC) and Area Under the ROC (AUC) curves including Kaggle National Data Science Bowl 2017 1st-place solution (Kaggle Top1) on our validation and testing data, image only learning, clinical information learning only, and proposed co-learning method. The classification confusion matrices, with class prediction thresholds selected based on ROC curves, are presented in Figure 7 as well. The proposed Co-learning model achieved superior performance on both validation and testing data compared with the baseline methods (e.g. the Co-learning model improves the AUC from 0.687 (image-only) and 0.635 (clinical-only) to 0.787).

## 4. CONCLUSION AND DISCUSSION

In this study, we propose a co-learning pipeline that integrates image features from 3D CT images and clinical demographics for lung cancer prediction. A 3D attention-based Convolutional Neural Net (CNN) was proposed to train on the CT scans and a random forest classifier was implemented to combine image features and clinical evidences for final lung cancer benign/malignant classification. The proposed co-learning strategy achieved superior performance compared with image only, clinical only, and Kaggle Top1 methods as seen from Figure 7. The major limitation of the work is the limited number of training scans, which is relatively small for deep learning. In the future, after acquiring more CT image scans with more detailed clinical information (blood, gene, phenotype etc.), we anticipate that the proposed method would achieve better performance.

## 5. ACKNOWLEGEMENTS

This research was supported by NSF CAREER 1452485, NIH grants 5R21EY024036, 1R21NS064534, 1R01EB017230 (Landman), 1R03EB012461 (Landman). This research was conducted with the support from Intramural Research Program, National Institute on Aging, NIH. This study was supported in part by a UO1 CA196405 to Massion. This study was in part using the resources of the Advanced Computing Center for Research and Education (ACCRE) at Vanderbilt University, Nashville, TN. This project was supported in part by ViSE/VICTR VR3029 and the National Center for Research Resources, Grant UL1 RR024975-01, and is now at the National Center for Advancing Translational Sciences, Grant 2 UL1 TR000445-06. We gratefully acknowledge the support of NVIDIA Corporation with the donation of the Titan X Pascal GPU used for this research. The imaging dataset(s) used for the analysis described were obtained from ImageVU, a research resource supported by the VICTR CTSA award (ULTR000445 from NCATS/NIH), Vanderbilt University Medical Center institutional funding and Patient-Centered Outcomes Research Institute (PCORI; contract CDRN-1306-04869).